\pdfoutput=1

\documentclass[11pt]{article}

\usepackage[]{acl}

\usepackage{times}
\usepackage{latexsym}

\usepackage[T1]{fontenc}

\usepackage[utf8]{inputenc}

\usepackage{microtype}

\usepackage{inconsolata}
\usepackage{enumitem}
\usepackage{adjustbox}
\usepackage{multirow}
\usepackage{caption}
\usepackage{amsmath}
\usepackage{graphicx}
\usepackage{booktabs}
\usepackage{mathtools}

\newcommand{\squishlist}{
 \begin{list}{$\bullet$}
  { \setlength{\itemsep}{0pt}
     \setlength{\parsep}{3pt}
     \setlength{\topsep}{3pt}
     \setlength{\partopsep}{0pt}
     \setlength{\leftmargin}{1.5em}
     \setlength{\labelwidth}{1em}
     \setlength{\labelsep}{0.5em} } }
\newcommand{\squishend}{
     \end{list}}

\title{TempoFormer: A Transformer for Temporally-aware Representations in Change Detection}

\author{
\textbf{Talia Tseriotou$^{1}$, Adam Tsakalidis$^{1,2}$}\\
\textbf{Maria Liakata$^{1,2}$}\\
       $^1$Queen Mary University of London, $^2$The Alan Turing Institute\\
      \tt  \{t.tseriotou,m.liakata\}@qmul.ac.uk}
\begin{document}
\maketitle

\begin{abstract}

Dynamic representation learning plays a pivotal role in understanding the evolution of linguistic content over time. On this front both context and time dynamics as well as their interplay are of prime importance. 
Current approaches model context via pre-trained representations, which are typically temporally agnostic. Previous work on modelling context and temporal dynamics has used recurrent methods, which are slow and prone to overfitting.
Here we introduce TempoFormer, the first task-agnostic transformer-based and temporally-aware model for dynamic representation learning. Our approach is jointly trained on inter and intra context dynamics and introduces a novel temporal variation of rotary positional embeddings. The architecture is flexible and can be used as the temporal representation foundation of other models or applied to different transformer-based architectures. 
We show new SOTA performance on three different real-time change detection tasks.

\end{abstract}

\section{Introduction}

\begin{figure}[h]
\centering
\includegraphics[width=.76\linewidth]{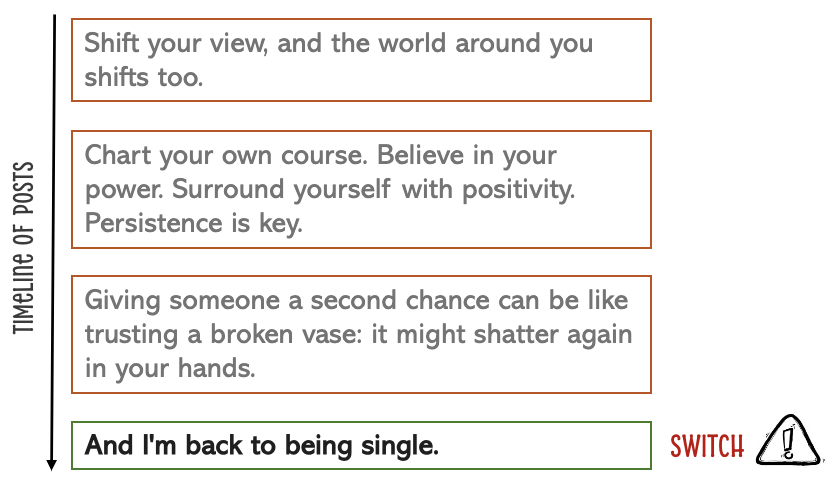}
\caption{
Paraphrased example from the task of identifying moments of change in individuals' mood \citep{tsakalidis2022identifying}. Here, the historical content (light grey) provides important linguistic context towards identifying a \textit{Switch}, a sudden mood shift from positive to negative, in the user's behaviour at the last post (black).}
\label{fig:context_example}
\end{figure}

Linguistic data sequences are generated continuously over time in the form of social media posts, written conversations or documents that keep evolving (e.g. through regular updates). While a large body of work has been devoted to assessing textual units or sub-sequences in isolation -- i.e. in emotion classification \citep{alhuzali2021spanemo}, ICD coding \citep{yuan2022code}, task-specific dialogue generation \citep{brown2024generation}, irony and sarcasm detection \cite{potamias2020transformer} -- such approaches leave significant historical (often timestamped) context unused. 
Fig.~\ref{fig:context_example} provides an example from the task of identifying mood changes through users' online content, where the last post in isolation cannot denote if there has been a \textit{`Switch'} in the user's mood -- the historical content provides important context for the user's originally positive mood, enhancing the signal for a negative switch in their behaviour.

\noindent\textbf{Dynamic representation learning} approaches aim to tackle this challenge. Dynamic word embedding methods have been studied in the context of semantic change detection \citep{bamler2017dynamic, rosenfeld2018deep}. 
While changes in this context occur over long time periods, dynamic representation learning has been explored in other more temporally fine-grained tasks such as event detection \citep{yan2019event, yang2019exploring, lai2020event}, 
fake news detection \citep{vaibhav2019sentence, raza2022fake, kaliyar2021fakebert} and mental health condition detection \citep{sawhney2021suicide, tsakalidis2022overview, tseriotou2023sequential}. Such temporally fine-grained tasks significantly differ from semantic change detection approaches: not only on the temporal granularity aspect, but crucially with respect to event timeline length, irregularities in change frequency, annotation requirements and problem formulation. Therefore the adaptation of methodologies between the various sets of temporal change detection categories is at best challenging. Correspondingly, fine-grained dynamic representation learning research remains also largely task or even dataset specific.

\noindent\textbf{Transformer-based injection}. The above mentioned approaches have relied on either pre-trained contextualised representations or transformer-based model layers \citep{devlin2018bert, liu2019roberta} to fine-tune representations before feeding them into RNN and CNN-like architectures as so far they had been shown to outperform transformer-based models \citep{ji2021does, gao2021limitations, tsakalidis2022identifying}. 
However, since LSTM-based systems tend to overfit small datasets, transformer-based methods that overcome this issue would be a preferable choice \citep{yu2020coupled}. Yet so far adapting layers on top of a transformer fails to strike the right balance between representation learning and task dynamics \citep{li2022multimodal,ng2023modelling}.

\noindent\textbf{Temporal modelling}. 
Although integration of time in language models has been explored for temporal adaption \citep{rottger2021temporal} in semantic change detection, \citep{rosin2022temporal,wang2023bitimebert} there is not yet work that explores the abilities of transformers to model temporally distant textual sequences (streams). Recently LLMs have been shown to fall short in terms of temporal reasoning \citep{jain2023language, wallat2024temporal}, especially in event-event temporal reasoning \citep{chu2023timebench}. Here we make the following contributions:

\squishlist
\item We present a novel, temporally-aware BERT-based model (\textbf{`TempoFormer'})\footnote{\url{https://github.com/ttseriotou/tempoformer}} that models streams of chronologically ordered textual information accounting for their temporal distance. 
TempoFomer is the first such model to directly modify the transformer architecture, doing so in a flexible and task-agnostic manner.
    \item We transform rotary position embeddings into rotary temporal embeddings that measure the temporal distance of sequential data points.
    \item 
    Contrary to prior work reliant on pre-trained contextual embeddings, we allow for adaptation of transformers towards the domain and the temporal aspects of a dataset. We show that TempoFormer can be used as the foundation in more complex architectures (e.g. involving recurrence), striking the right balance between modelling a post/utterance (context-aware) and the timeline-level dynamics. Moreover the TempoFormer upper layers are flexible and can be applied 
    in different Transformer-based architectures.
    \item We show SOTA performance on 3 change detection NLP tasks (longitudinal stance switch, identifying mood changes and identifying conversation derailment).
\squishend

\section{Related Work}

\noindent \textbf{Context-aware Sequential Models:}
Numerous social media related tasks such as rumour detection rely on chronologically ordered conversation threads \cite{ma2020attention,lin2021rumor,ma2020debunking}. 
Moreover \citet{ng2023modelling} have shown lift in performance  when using the full context of medical notes, rather than the discharge summary alone, for ICD coding. However context-aware sequential models have so far relied on recurrent networks or hierarchical attention \citep{li2020multi, ma2020attention,tsakalidis2022overview} without exploring the dynamics between sentence level and stream level representations. \\
\noindent \textbf{Longitudinal Modelling and Change Detection:} 
In addition to the importance of the linguistic stream, longitudinal tasks rely on temporal dynamics to asses progression 
and identify changes over time. 
In the case of (a) \textit{identifying changes in user mood} \citep{tsakalidis2022identifying, tsakalidis2022overview, tseriotou2023sequential, hills2024exciting} and  suicidal ideation through social media \citep{sawhney2021phase} change is relative to the temporal evolution of users' mood over time and approaches have relied mostly on recurrence and on utterance-level pretrained language model (PLM) representations. \citet{tseriotou2024sig} introduced a longitudinal variation of (b) \textit{stance detection} \cite{yang2022weakly, kumar2019tree} for detecting shifts (changes) in the public opinion towards an online rumour. They used Sentence-BERT \citep{reimers2019sentence} representations with integration of path signatures \citep{lyons1998differential} in recurrence. For (c) \textit{conversation topic derailment}, 
previous work has relied on fine-tuning transformer-based models \citep{konigari2021topic}, providing extended context in their input \citep{kementchedjhieva2021dynamic} or applying recurrence over the utterance \citep{zhang2019generic} and context stream \citep{chang2019trouble}. 
In this work we integrate stream dynamics directly into the transformer and show the flexibility of our approach as the foundation of different longitudinal models.


\noindent \textbf{Temporal Language Modelling: }
Many of the above tasks involve timestamps, which can enhance change detection through temporal dynamics. 
However, little research in NLP leverages time intervals and those who do assume equidistant time intervals between events \citep{ma2020debunking, tsakalidis2020sequential}. Other work on temporal modelling has relied on hand crafted periodic task-specific time features \citep{kwon2013prominent}, concatenation of timestamp with linguistic representations \citep{tseriotou2023sequential, tseriotou2024sig} or Hawkes temporal point process applied on top of recurrence \citep{guo2019personalized,hills2024exciting}. These approaches applied on top of LM representations miss the opportunity of training representations informed by temporal dynamics. 
Additionally, transformer-based models lack temporal sensitivity \citep{lazaridou2021mind, loureiro2022timelms}. \citet{rosin2022temporal} has conditioned attention weights on time, while \citet{rosin2022time, wang2023bitimebert} concatenated time tokens to text sequences. Although these methods create time-specific contextualised embeddings, they utilise absolute points in time rather than leveraging the temporal distance between units of textual information, important for context-aware and longitudinal tasks. Here we adapt the transformer attention mechanism to cater for 
the relative temporal aspect (\S \ref{sec:romha}). 

\noindent \textbf{Hierarchical Models:}
Long content modelling approaches have leveraged transformer or attention-based blocks hierarchically on long documents, on input chunks/sentences and then on the sequence of such chunks \citep{zhang2019hibert, pappagari2019hierarchical,wu2021hi,li2023recurrent}. This produces chunk-level summary embeddings, which preserve both the local and global aspects of contextualised representations. 
Here we leverage such local and global context dynamics to more efficiently model linguistic streams. 

\section{Methodology}

Here we introduce the TempoFormer architecture. We first provide the problem formulation (\S \ref{sec:formulation}), followed by model overview (\S \ref{sec:tempoformer_overview}) and then discuss the various model components (\S \ref{sec:post_encoding}-\ref{sec:tuning}).

\subsection{Problem Formulation} \label{sec:formulation}
A fundamental concept underpinning longitudinal tasks is that of \textit{timelines}, $P$, defined as chronologically ordered units of information between two dates \citep{tsakalidis2022identifying}, here either in the form of a sequence of users' posts, a conversation or an online thread. Specifically the $c$-th timeline, $P^{c}$, consists of a series of posts\footnote{We use terms \textit{posts} and \textit{utterances} interchangeably as the exact nature of the textual unit depends on the specific task.}, $u_{i}$, each with a corresponding timestamp, $t_{i}$. $P^{c}=[\{u_{0},t_{0}\}, \{u_{1},t_{1}\},...,\{u_{N-1},t_{N-1}\}]$. The length of the timeline, $N$, can vary.
We formulate the problem of assessing textual units in a timeline as early-stage, real-time classification, following \citet{tseriotou2023sequential}. We map each \textit{timeline} into $N$ training samples, that we call \textit{streams}. Each \textit{stream} contains a predefined window, $w$, of the most recent posts and a label for the most recent post: $([\{u_{i-w+1},t_{i-w+1}\},... \{u_{i-1},t_{i-1}\},\{u_{i},t_{i}\}], l_{i})$. 

\subsection{TempoFormer Overview}
\label{sec:tempoformer_overview}

\begin{figure*}[h]
\centering
\captionsetup{justification=justified}
\includegraphics[width=.85\linewidth]{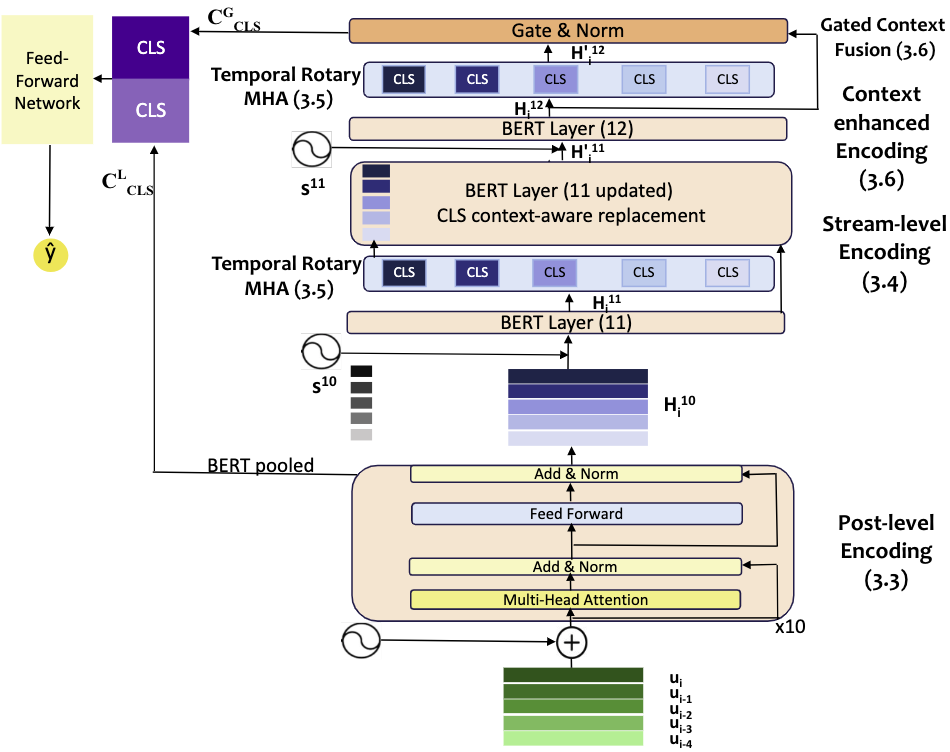}
\caption{TempoFormer Architecture on 5-post window.}
\label{fig:tempoformer}
\end{figure*}

Fig.~\ref{fig:tempoformer} provides an overview of \textbf{TempoFormer}. Its hierarchical architecture consists of three main modules, temporally-aware enhancement in multi-head attention to model the temporal distance between posts and a classification head. The modules are: \textbf{post-level (local) encoding} (\S \ref{sec:post_encoding}) -- obtaining word-level representation of each post using BERT's first 10 layers; \textbf{stream (global) encoding} (\S \ref{sec:stream_encoding}) -- modelling the sequential and temporal interactions between posts; and \textbf{context-enhanced encoding} (\S \ref{sec:context_encoding}) -- fusing stream-awareness in post-level representations to make them context-aware. 

\subsection{Post-level Encoding (Local)} \label{sec:post_encoding}
Each training instance is a stream consisting of the current post and its recent history,  alongside corresponding timestamps: $[\{\mathbf{u}_{i-w+1},t_{i-w+1}\},... \{\mathbf{u}_{i-1},t_{i-1}\},\{\mathbf{u}_{i},t_{i}\}]$, with a total of $w$ posts in a stream. Timestamps are ignored at this stage. This stream of posts is converted into a stream, $e$, of word-level embeddings of word sequence length $K$ via the word and position embedding layer of BERT: $[\{\mathbf{e}_{1,i-w+1},\mathbf{e}_{2,i-w+1}...,\mathbf{e}_{K,i-w+1}\},...$
$\{\mathbf{e}_{1,i},\mathbf{e}_{2,i}...,\mathbf{e}_{K,i}\}]$. Specifically, in this module, the posts in each \textit{stream} pass without post-post interactions via the first 10 BERT layers, resulting in hidden word-level representations for each post. Note that since a post is part of multiple streams through their window, it will pass through the BERT layers as part of each corresponding stream. For each post $j$ (belonging to a stream $q$), the word-level representations from the $z$-th Transformer layer are denoted as: $\mathbf{H}^z_{j_q}$ 
Therefore at the 10-th layer we reconstruct the stream and form local stream representation: $[\mathbf{H}^{10}_{i-w+1},...,\mathbf{H}^{10}_{i}]$.

\subsection{Stream-level Encoding (Global)} \label{sec:stream_encoding}
Inspired by \citet{wu2021hi}, who model long documents hierarchically by stacking transformer-based layers of sentence, document and document-aware embeddings, we build stream and context-enhanced layers on top of post-level representations. At the stream encoding layer, we capture inter-stream dynamics. Stream-level position embeddings (PE), $\mathbf{s}^{10}$, added after the 10-th layer, encode post order within the stream. By then passing the word-level stream PE representations 
to another BERT layer, we obtain word-level sequence-aware updated hidden representations $[\mathbf{H}^{11}_{1,i-w+1}$, ...,$\mathbf{H}^{11}_{1,i}]$.

Next, we obtain the order-aware \texttt{[CLS]} token from the stream and apply \textbf{Temporal Rotary Multi-head Attention} (MHA), a proposed variation of RoFormer \citep{su2024roformer}, which accounts for the \textit{temporal} rather than the \textit{sequential} distance between posts (see \S\ref{sec:romha}). These context-aware, temporally-enhanced tokens are fed back to replace the respective \texttt{[CLS]} tokens in the  hidden representations from the previous BERT layer, resulting in $[\mathbf{H}^{'11}_{1,i-w+1}$, ...,$\mathbf{H}^{'11}_{1,i}]$. This enables the propagation of the learnt stream embeddings to the post-level.

\subsection{Temporal Rotary Multi-Head Attention} \label{sec:romha}
BERT  relies on positional embeddings to meaningfully encode the sequential order of words which are then fused via self-attention. Such embeddings are absolute (position-specific) and lack a relative sense. \citet{su2024roformer} proposed the Rotary Position Embeddings (RoPE) that incorporate
the \textit{relative} position between tokens within self-attention. Besides flexibility (in terms of sequence length generalisability), this introduces in the formulation intuitive inter-token dependency, which decays with increasing token distance. Given the  attention formulation $\textrm{Attn}(\mathbf{Q},\mathbf{K},\mathbf{V})_m= \dfrac{\sum^{N}_{n=1} (\textrm{exp}(\mathbf{q}^T_m \mathbf{k}_n/\sqrt{d}))\mathbf{v}_n}{\sum^{N}_{n=1} (\textrm{exp}(\mathbf{q}^T_m \mathbf{k}_n/\sqrt{d}))}$, where $m$/$n$ denote the query/key positions, 
 after applying RoPE self-attention, the  $\mathbf{q}^T_m \mathbf{k}_n$ becomes:
\begin{equation} \label{eq:rotary}
\resizebox{.85\hsize}{!}{$
        \mathbf{q}^T_m \mathbf{k}_n=(R^d_{\theta,m}\mathbf{q}_m)^T(R^d_{\theta,n}\mathbf{k}_n)
        =\mathbf{q}^T_m R^d_{\theta,n-m} \mathbf{k}_n
        $},
\end{equation}
\noindent where $R^d_{\theta,m}$ is the rotary matrix with $d$ embedding dimensions and the following formulation:
\begin{equation*}
\resizebox{.9\hsize}{!}{$
    R^d_{\theta,m}=
  \begin{pmatrix*}[c]
   \textrm{cos}(m\theta_1) & -\textrm{sin}(m\theta_1)  & 0 &0 \\
   \textrm{sin}(m\theta_1) & \textrm{cos}(m\theta_1)  & 0 &0 \\ 
    \vdots & \ddots  & \ddots & \vdots \\ 
   0 & 0  & \textrm{cos}(m\theta_{d/2}) & -\textrm{sin}(m\theta_{d/2})\\ 
   0 & 0  & \textrm{sin}(m\theta_{d/2}) & \textrm{cos}(m\theta_{d/2}) \\    
    \end{pmatrix*}
    $}
\end{equation*}
\noindent where $\theta_i = 10000^{-2(i-1)/d} i \in [1, 2, ..., d/2]$. 
The rotary matrix incorporates the relative position information through rotation of $q$ and $k$ based on their position in the sequence. The dot product decreases as the tokens move further apart. In Eq.~\ref{eq:rotary}, the formulation results in the relative position $(m-n)$, so the rotation between the 6-th and the 3-rd tokens is the same as between the 7-th and the 4-th ones.

Here, in order to model the temporal dynamics, we propose a novel variation of Eq.~\ref{eq:rotary}, 
named \textbf{Temporal Rotary Multi-head Attention}, making use of the relative position property. Instead of $R^d_{\theta,n-m}$, we reformulate the rotary matrix to model the \textit{temporal},  rather than the positional differences, $R^d_{\theta,\mathbf{t_n-t_m}}$. We employ it on the stream-level using the \texttt{[CLS]} tokens to capture the stream global context through both the temporal and linguistic dynamics. The developed layer includes solely self-attention without the need for feed-forward and normalisation layers. In practice, since we measure time in seconds, we log-transform time in order to remove task dependencies on the scale of temporal propagation, to account for stream non-linearities and to alleviate exclusion of temporal outliers.

\subsection{Context-enhanced Encoding} \label{sec:context_encoding}
Literature has shown the effectiveness of enhancing word-level representations hierarchically through context-level learnt dynamics \citep{zheng2020towards,wu2021hi,ng2023modelling}. To this effect we introduce a second-layer of stream-level position embeddings, $\mathbf{s}^{11}$, to re-instate the absolute sequence position of each post for context-enhanced modelling. These are fed into a global context-aware layer, essentially a word-level transformer layer.  Since the \texttt{[CLS]} tokens of each post are stream-aware, they contextualise the token-level representations based on the temporal and global learnt dynamics, obtaining: $[\mathbf{H}^{12}_{1,i-w+1},...,\mathbf{H}^{12}_{1,i}]$. To fully model the stream dynamics given the now context-enhanced \texttt{[CLS]} tokens, we employ a last layer of Temporal Rotary MHA resulting in $[\mathbf{H}^{'12}_{1,i-w+1},...,\mathbf{H}^{'12}_{1,i}]$. Lastly, we adapt the \textit{Gated Context Fusion} (Gate\&Norm) mechanism by \citet{zheng2020towards} to fuse both the utterance word-level informed ($\mathbf{H}^{12}_{CLS}$) and the stream utterance-level informed ($\mathbf{H}^{'12}_{CLS}$) \texttt{[CLS]} tokens through element-wise multiplication $\odot$:
\vspace{-.2cm}
\begin{equation*}
        \resizebox{.5\hsize}{!}{$\mathbf{g} = \sigma(W_g[\mathbf{H}^{12}_{CLS}; \mathbf{H}^{'12}_{CLS}])$} \\
\end{equation*}
\vspace{-1.5cm}

\begin{equation*}
        \resizebox{.9\hsize}{!}{$\mathbf{C}^G_{CLS} = \textrm{LayerNorm}[(1-\mathbf{g})\odot {H}^{12}_{CLS} + \mathbf{g} \odot {H}^{'12}_{CLS}]$}
\end{equation*}

\subsection{Network Fine-Tuning} \label{sec:tuning}

Although the proposed architecture can in principle be applied to any Transformer-based model, we select BERT \citep{devlin2018bert} as the foundation model and initialise all word-level weights. Literature on longitudinal context-aware classification has shown the importance of efficiently combining the current utterance with historical information \citep{sawhney2020time,sawhney2021phase,tseriotou2023sequential}. We thus concatenate the \textit{local stream-agnostic \texttt{[CLS]}} token of the current utterance from the 10-th layer, $\mathbf{C}^L_{CLS}$, (obtained through typical BERT Pooling) with the obtained \textit{global stream-enhanced} \texttt{[CLS]}, $\mathbf{C}^G_{CLS}$ (Fig.~\ref{fig:tempoformer}). This final representation is fed through two fully connected layers with ReLU activation and dropout \citep{srivastava2014dropout}. The architecture is fine-tuned for each classification task (\S \ref{sec:tasks}) using alpha-weighted focal loss \citep{lin2017focal}, to assign more importance to minority classes and alleviate class imbalance.

\section{Experiments} \label{sec:tasks}

\subsection{Tasks and Datasets}
We test our model on three different longitudinal change detection classification tasks of different temporal granularity: 1) \textbf{Stance Switch Detection} -- identification of switches in overall user stance around a social media claim, 2) \textbf{Moments of Change (MoC)} -- identification of mood changes through users' online posts and 3) \textbf{Conversation Topic Shift} -- conversation diversion identification. We adopt a real-time prediction formulation (see  \S\ref{sec:formulation}) to assess system ability to perform early change detection in real-world scenarios. Table \ref{tab:statistics}  provides detailed statistics for each dataset, showing the different degrees of temporal granularity and dataset specifics. More details on data splits and stream examples are provided in Appendix \ref{app:sec:datasets} and \ref{app:sec:examplesdataset} respectively.\\

\noindent \textbf{Stance Switch Detection}: Introduced by \citet{tseriotou2024sig} takes a sequence of chronologically ordered Twitter conversations about a rumourous claim related to a newsworthy event, to detect switches in the overall user stance. Conversations are converted from a tree structure into a chronologically ordered linear list (timeline). We use the \textbf{LRS} dataset from \citet{tseriotou2024sig} based on RumourEval-2017 \citep{gorrell2019semeval}, and convert the original stance labels (supporting/denying/questioning/commenting) with respect to the root claim into two categories of 
\textit{Sw}: (switch) a shift in the number of opposing (denying/questioning) vs supporting posts and \textit{N-Sw}: absence of switch or cases where the numbers of supporting and opposing posts are equal. Each post is accompanied by its timestamp.

\noindent \textbf{Moments of Change (MoC)}: Introduced by \citet{tsakalidis2022identifying} takes a sequence of chronologically ordered posts shared by an online social media user, and classifies each post according to the behavioural change of the user as one of: \textit{IS}- (switch) sudden mood shift
from positive to negative (or vice versa); \textit{IE}- (escalation) gradual mood progression from neutral or positive/negative to more positive/negative; and \textit{O}- no mood change. We use the \textbf{TalkLife} dataset from \citet{tsakalidis2022identifying} containing such annotated timelines where each post is timestamped.

\noindent \textbf{Conversation Topic Shift}: Given a corpus of open-domain conversations between humans, this binary classification task identifies whether each utterance falls under the main conversation topic or if it has derailed from it. We use the \textbf{Topic Shift-MI} (Mixed-Initiative) dataset \citep{konigari2021topic} annotated on a subset of the Switch-board dataset~\citep{godfrey1992switchboard, calhoun2010nxt}. This dataset has a single but varying major topic for each conversation. The two classes are \textit{M}: (major) utterance belongs to the main topic and \textit{R}: (rest) utterance pertains to a minor topic or is off-topic. Here conversations are not timestamped.

\begin{table}[ht]
\begin{adjustbox}{max width=\linewidth}
\begin{tabular}{|l|l|l|l|}
\hline
\multicolumn{1}{|c|}{\multirow{1}{*}{\textbf{Dataset}}} &    \multicolumn{1}{c|}{\textbf{LRS}}         & \multicolumn{1}{c|}{\textbf{TalkLife}}  
 & \multicolumn{1}{c|}{\textbf{Topic Shift MI}}     
   \\ \cline{1-4} 
{\#  Data Points} & \multicolumn{1}{c|}{5,568} &\multicolumn{1}{c|}{18,604} & \multicolumn{1}{c|}{12,536} \\ \hline
{\#  Timelines} & \multicolumn{1}{c|}{325} &\multicolumn{1}{c|}{500} & \multicolumn{1}{c|}{74} \\ \hline
{Mean (median)} & \multicolumn{1}{c|}{17.1 } &\multicolumn{1}{c|}{37.2} & \multicolumn{1}{c|}{169.4} \\ 
{Timeline Length in Posts} & \multicolumn{1}{c|}{(13)} &\multicolumn{1}{c|}{(30)} & \multicolumn{1}{c|}{(153.5)} \\ \hline
{Mean (median)}& \multicolumn{1}{c|}{1h 26m 40s}             & \multicolumn{1}{c|}{6h 51m 11s}     
& \multicolumn{1}{c|}{-}  \\ 
{Time inbetween Posts}& \multicolumn{1}{c|}{1m 39s}             & \multicolumn{1}{c|}{59m 38s}     
& \multicolumn{1}{c|}{(-)}  \\ \hline
{Mean (median)}  &  \multicolumn{1}{c|}{6.5}            & \multicolumn{1}{c|}{IS:1.8, IE:4.0}     
& \multicolumn{1}{c|}{60.5}\\ 
{\# Minority Events/Timeline}  &  \multicolumn{1}{c|}{(0)}            & \multicolumn{1}{c|}{(IS:1, IE:1)}     
& \multicolumn{1}{c|}{(51.5)}\\ \hline
\end{tabular}
\end{adjustbox}
\caption{Statistics of Datasets.}
\label{tab:statistics}
\end{table}

\subsection{Baselines and Experimental Setup}
We select classification baselines that are both \textit{post-level} (current post only) and \textit{stream-level} (recent window of chronologically ordered posts including the current post, see \S\ref{sec:formulation}). To account for the class imbalance, we use focal loss \citep{lin2017focal} for all the fine-tuned models.


\noindent \textit{\underline{Post-level}}:

\noindent \textbf{Random}: post classification based on probabilities of class distributions.

\noindent \textbf{BERT/RoBERTa}: BERT \citep{devlin2018bert} or RoBERTa \citep{liu2019roberta} fine-tuned.

\noindent \textbf{Llama2-7B-U (5/10-shot)}: In-context learning with Llama2-7B-chat-hf LLM \citep{touvron2023llama} using a crafted prompt for each dataset. Experiments on both 5 and 10 few-shot examples were randomly sampled to reflect the distribution of the dataset, following \citet{min2022rethinking}.

\noindent \textbf{MistralInst2-7B-U(5/10-shot)}: Same setting as for Llama2-7B-U, using the Mistral-
7B-Instruct-v0.2~\citep{jiang2023mistral} LLM.

\noindent \textit{\underline{Stream-level}}:

\noindent\textbf{FFN History}: Feed-forward network of 2 hidden layers on the concatenation of SBERT \citep{reimers2019sentence} embeddings a) of the current post and b) averaged over the window posts.

\noindent\textbf{SWNU} \citep{tseriotou2023sequential}: Expanding windows of path signatures applied over learnable dimensionally-reduced data streams of SBERT representations and time and fed into a BiLSTM to model the information progression.

\noindent\textbf{Seq-Sig-Net} \citep{tseriotou2023sequential}: Sequential BiLSTM Network of SWNU units that capture long-term dependencies concatenated with the current post's SBERT representation.

\noindent\textbf{BiLSTM}: Bidirectional single-layer recurrent network applied on the stream of SBERT embeddings.

\noindent \textbf{Llama2-7B-S (5-shot)}: 5-shot in-context learning following the same set up as in Llama2-7B-U but including the recent history of window 5 in each shot (for context) instead of only the current post.

\noindent \textbf{MistralInst2-7B-S (5/10-shot)}: Same 5 and 10 few-shot setting as for Llama2-7B-S, using the Mistral-7B-Instruct-v0.2 LLM.


\noindent \textbf{Evaluation:} In line with published literature we report F1 scores for model performance, per class and macro-averaged. For each dataset we perform 5-fold cross validation with train/dev/test sets consisting of different timelines. We run and report the performance of each model on the exact same four random seeds (0,1,12,123) and report the average result (as well as the standard deviation on macro-average) on the test set. Appendix \ref{sec:app:params} provides information about implementation details and hyperparameter search.

\section{Results and Discussion}

\subsection{Comparison against baselines}
We present results for TempoFormer and baselines in Table \ref{tab:mainresults_presentation}. TempoFormer is the most performant in all three tasks based on macro-averaged F1. We note that recurrent models based on pre-trained BERT representations (BiLSTM for LRS and Topic Shift MI and Seq-Sig-Net for TalkLife), ranked second best. The latter models have been the SOTA for these datasets \citep{tseriotou2023sequential, tseriotou2024sig}. While the datasets are of different sizes, temporal characteristics, timeline length and change event distribution (see Table \ref{tab:statistics}), TempoFormer retains its high performance, showcasing its generalisability for real-time change detection. Importantly, our model has the highest F1 for all minority classes, with the exception of Topic Shift MI, where other baselines have higher class-specific F1 scores for M but much lower F1 for R. Since TempoFormer operates on a contextual window of recent posts we select the appropriate window for each stream based on a window analysis, reported in \S\ref{sec:window}.

\begin{table*}
\begin{adjustbox}{max width=.9\textwidth,center}
\begin{tabular}{{|c|l|rrc|rrrc|rrc|}}
\hline
 & & \multicolumn{3}{|c|}{\textbf{LRS}}
 & \multicolumn{4}{|c|}{\textbf{TalkLife}} & \multicolumn{3}{|c|}{\textbf{Topic Shift MI}} \\
 
&\textbf{Model}  & N-Sw & Sw & macro-avg     & IE & IS & O&macro-avg      & M & R & macro-avg \\\cline{1-12}
\parbox[t]{2mm}{\multirow{7}{*}{\rotatebox[origin=c]{90}{Post-level}}}& Random& 61.4 &37.5  & 49.5$^{\pm0.510}$&11.2 & 4.5 & 84.4 & 33.4$^{\pm0.080}$ & 35.9&  63.9& 49.9$^{\pm0.332}$ \\
&Llama2-7B-U (5-shot)& 22.4 & 50.6&  36.5$^{\pm0.000}$&  10.1 &7.5&31.9&16.5$^{\pm0.000}$& 46.6&  45.4&46.0$^{\pm0.000}$ \\
  &  MistralInst2-7B-U (5-shot)& 71.4 &28.0&49.7$^{\pm0.000}$& 23.3&  4.1& 67.8& 31.7$^{\pm0.000}$&   46.4&  44.6& 45.5$^{\pm0.000}$\\
& Llama2-7B-U (10-shot) & 8.8 & 52.5 & 30.7$^{\pm0.000}$ & 12.8 & 6.2 &  31.3 &16.7$^{\pm0.000}$ & 48.5& 39.5 & 44.0$^{\pm0.000}$\\
& MistralInst2-7B-U (10-shot) & 71.2 & 30.5&50.8$^{\pm0.000}$  & 27.6 &3.5  &72.1&34.4$^{\pm0.000}$ &42.6  &55.7  & 49.1$^{\pm0.000}$ \\
& BERT& 69.0 & 45.3 & 57.1$^{\pm0.995}$& 43.9 & 28.1 & 86.8& 52.9$^{\pm0.140}$ & 36.0& 70.0 & 53.0$^{\pm0.186}$ \\
& RoBERTa&68.2  & 46.4 & 57.3$^{\pm1.280}$& 46.3 & 30.4 & 86.6& 54.4$^{\pm0.321}$ &34.5 & 70.2 & 52.4$^{\pm0.266}$\\
\midrule
\bottomrule

\parbox[t]{2mm}{\multirow{8}{*}{\rotatebox[origin=c]{90}{Stream-level}}}& FFN History &71.6 & 52.8& 62.2$^{\pm0.915}$ & 45.4 &  27.1&88.0 &  53.5$^{\pm0.372}$& 39.4&  70.1& 54.8$^{\pm0.448}$\\
 & SWNU & 75.5 & 55.5 & 65.5$^{\pm0.715}$& 48.0& 29.3& \textbf{89.5}& 55.6$^{\pm0.461}$& 38.7& 66.0& 52.3$^{\pm0.749}$\\
& Seq-Sig-Net  & 74.7&  58.9& 66.8$^{\pm0.487}$& 48.4& 30.2& \textbf{89.5}&56.0$^{\pm0.219}$ &37.4 & 66.7&52.1$^{\pm0.977}$ \\
 & BiLSTM & 75.0 &60.7& 67.8$^{\pm1.400}$& 46.1& 27.0 &89.2  &54.1$^{\pm0.113}$  & 37.8 & \textbf{73.8} & 55.8$^{\pm0.672}$\\
& Llama2-7B-S (5-shot) & 2.2 & 50.2& 26.2$^{\pm0.000}$ & 15.5 & 7.6 & 24.2 & 15.7$^{\pm0.000}$ & \textbf{52.6}& 1.3 & 27.0$^{\pm0.000}$\\
& MistralInst2-7B-S (5-shot)  &58.3 & 50.2 &54.3$^{\pm0.000}$ & 22.0 &  4.6 & 70.0 &  32.2$^{\pm0.000}$& 42.3& 57.3 & 49.8$^{\pm0.000}$\\
& MistralInst2-7B-S (10-shot)  &54.4 & 51.8 &53.1$^{\pm0.000}$ & 23.4 & 3.5 & 74.9 &  33.9$^{\pm0.000}$& 37.8&  63.7& 50.8$^{\pm0.000}$\\
& TempoFormer (ours) & \textbf{75.9}& \textbf{62.0}& \textbf{68.9}$^{\pm1.409}$&  \textbf{50.0}& \textbf{32.4} &88.8 & \textbf{57.1}$^{\pm0.352}$ & 41.6& 70.7 &\textbf{56.1}$^{\pm0.463}$  \\
\bottomrule
\end{tabular}
\end{adjustbox}
\caption{(\textbf{Best}) F1-scores across all tasks. Stream-level models are applied on the optimal window, per dataset.}
\label{tab:mainresults_presentation}
\end{table*}

We distinguish baselines into post and stream-level ones, noticing that smaller fine-tuned Language Models, even as simple as an FFN, allowing for stream-level context, score consistently better than post-level ones - with the exception of RoBERTa for TalkLife. This consistent finding underscores the importance of developing contextually informed representations for change detection. Few-shot prompted LLMs have consistently lower performance  than smaller fine-tuned LMs, in line with reported poor performance of LLMs on temporal tasks~\citep{jain2023language, bian2023chatgpt}. For post-level, while Mistral's performance improved from 5 to 10-shot, it is still barely above the random baseline and significantly behind BERT and RoBERTa. For LRS and Topic Shift MI the stream-level 5 and 10-shot Mistral performance increases, but falls way short of BERT/RoBERTa and all the stream-level models, indicating that although sequential context is important it is not modelled appropriately with current LLMs. In line with \citep{wenzel2024temporal}, Llama2 suffers from generating responses outside the predefined classes, resulting in very low performance. TempoFormer demonstrates a generalisable architecture that enhances word-level post representations given the context, while modelling effectively the interplay between linguistic and temporal dynamics. 
\subsection{Window Length} \label{sec:window}

\begin{figure}
\centering
\includegraphics[width=.8\linewidth]{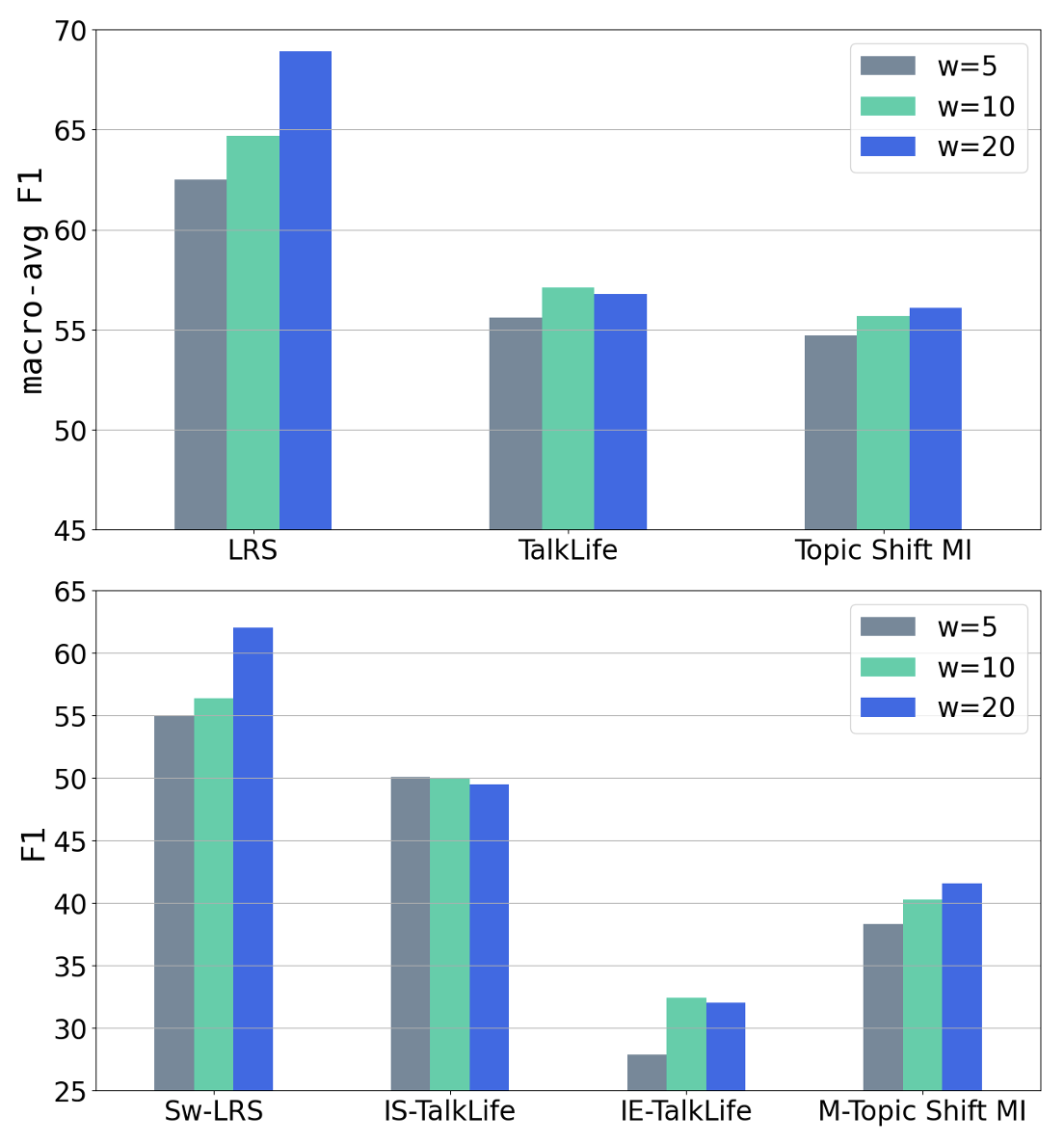}
\caption{TempoFormer Results for Different Contextual Window Sizes.}
\label{fig:window_plots}
\end{figure}

\noindent Since stream-based models operate on recent context, selecting appropriate contextual windows to include in the stream is important. Following \citet{tseriotou2024sig} we determine window selection based both on model performance and dataset characteristics. Fig. \ref{fig:window_plots} demonstrates TempoFormer’s F1 performance on windows of 5, 10 and 20 recent posts (see Table \ref{tab:window_presentation} for full results). While LRS and Topic Shift MI both benefit from the large window of 20 posts (blue) with clear performance gains overall and for the minority classes, TalkLife demonstrates better performance over a window of 10 (green). The optimal window findings for LRS and TalkLife are consistent with \citep{tseriotou2024sig}. 
These differences are attributed to dataset characteristics (Table \ref{tab:statistics}) and the mean number of change events in timelines, which need to be captured within the contextual windows. This analysis informs our stream-level experiments and at the same time demonstrates the flexibility of TempoFormer with respect to contextual window length. We recommend exploratory analysis according to dataset characteristics for appropriate window selection for new datasets. 


\subsection{Ablations Study}
In Table \ref{tab:ablation_table} we present an ablation study to assess the effect of each of TempoFormer's components.

\noindent \textbf{Temporal Rotary Multi-head
Attention (MHA)}: By using the vanilla sequential distance version of RoPE in Multi-head attention instead of the temporal one, for the timestamped datasets, we see a drop in performance. This showcases the advantage of modelling linguistic streams while accounting for their temporal dynamics and the success of temporally distant RoPE. The relatively small drop in performance is due to the secondary role of temporal dynamics compared to linguistic evolution in change detection tasks.

\noindent \textbf{RoPE MHA}: By further replacing the RoPE MHA with the vanilla version of MHA we see a significant drop in performance (in macro-avg): -2.9\% for LRS, -1.2\% for TalkLife and -0.6\% for Topic Shift MI, demonstrating the success of RoPE on its own. We postulate that this signifies the ability of RoPE to enable MHA integration in architectures without the need for normalisation and FFN in a full transformer layer.

\noindent \textbf{Stream embeddings}: Removing only the $s11$ embedding from the top layer results in performance drop, signifying the importance of re-integrating the absolute post position for context enhancement of word representations. Further ablating both of $s10$ and $s11$ embeddings from TempoFormer layers brings even more noticeable performance drops in all datasets, showcasing the overall significance of propagating sequence position information in building stream-aware and context-enhanced post embeddings. Topic Shift MI shows the largest drop of -1.4\% among all its ablated models. Since this dataset does not obtain sequential signal from temporal dynamics, it relies on stream embeddings to model the distance between consecutive posts.

\noindent \textbf{Gate\&Norm} operation updates the stream post-level \texttt{[CLS]} tokens with post word-level information, which is better informed by the word-level dynamics. This fuses together the word and stream dynamics in a gated learnable way. Large performance drops for all tasks when we ablate this component shows the importance of multi-level fusion.

\begin{table}
\begin{adjustbox}{max width=\linewidth,center}
\begin{tabular}{{|l|ccc|cccc|ccc|}}
\hline
& \multicolumn{3}{|c|}{\textbf{LRS}}
 & \multicolumn{4}{|c|}{\textbf{TalkLife}} & \multicolumn{3}{|c|}{\textbf{Topic Shift MI}} \\
 
\textbf{Models} & N-Sw & Sw & macro-avg     & IE & IS & O&macro-avg      & M & R & macro-avg \\\cline{1-11}

TempoFormer &\textbf{75.9}& \textbf{62.0}& \textbf{68.9} &\textbf{50.0} & \textbf{32.4} &88.8 & \textbf{57.1} &41.6 &70.7 &\textbf{56.1}\\

\hline

$\neg$Temporal RoPE & 75.5&\textbf{62.0}& 68.7& 49.3&  31.7& 88.7& 56.6 & - & - & -\\
\hline

$\neg$RoPE MHA & 74.1&57.9& 66.0& 48.0&31.5& 88.2& 55.9& 39.6&  \textbf{71.4}&55.5\\

\hline

$\neg$Stream embed. $s11$ & 75.7 & 60.1 & 67.9 & 49.7 & 32.1 & 88.9& 56.9& \textbf{43.7} & 68.2 & 55.9 \\

\hline

$\neg$Stream embed. $s10,s11$ &75.4 &59.0& 67.2& 49.4& 31.7 & \textbf{89.2}&  56.8& 38.9& 70.5 &54.7\\

\hline

$\neg$Gate\&Norm &74.5 & 61.3& 67.9& 49.8&  31.1& 88.7& 56.6 & 40.7& 69.6 &55.2\\

\bottomrule
\end{tabular}
\end{adjustbox}
\caption{Ablation Studies for TempoFormer based on F1 with one component ablated at a time for all datasets.}
\label{tab:ablation_table}
\end{table}

\subsection{The curious case of recurrence}
Since longitudinal and change detection models have so far heavily relied on recurrence-based architectures, we evaluate the effect of recurrence on models jointly trained for stream and post-level representations. To do so we adapt RoBERT \citep{pappagari2019hierarchical}, originally developed for long document classification, which applies recurrence over BERT's pooled \texttt{[CLS]} for each post. Here, we propose RoTempoFormer, a modification of RoBERT, that uses recurrence over TempoFormer's pooled \texttt{[CLS]} for each post. Both RoBERT and RoTempoFormer are stream-level, recurrence-based models. We report results in Table \ref{tab:recurrence_results}.

\begin{table}
\begin{adjustbox}{max width=\linewidth,center}
\begin{tabular}{{|c|ccc|cccc|ccc|}}
\hline
& \multicolumn{3}{|c|}{\textbf{LRS}}
 & \multicolumn{4}{|c|}{\textbf{TalkLife}} & \multicolumn{3}{|c|}{\textbf{Topic Shift MI}} \\
 
\textbf{model} & N-Sw & Sw & macro-avg     & IE & IS & O&macro-avg      & M & R & macro-avg \\\cline{1-11}

TempoFormer & 75.9 & 62.0 & 68.9$^{\pm1.409}$& \textbf{50.0} &\textbf{32.4} &\textbf{88.8} &\textbf{57.1}$^{\pm0.352}$ &\textbf{41.6} &70.7 &\textbf{56.1}$^{\pm0.463}$ \\

\midrule
\bottomrule

RoBERT & 75.8& 62.3& 69.0$^{\pm0.689}$& 36.7& 3.3& 88.4& 42.8$^{\pm0.565}$& 33.3& \textbf{75.7}& 54.5$^{\pm0.303}$\\
\midrule
\bottomrule

RoTempoFormer & \textbf{76.2}& \textbf{63.6}& \textbf{69.9}$^{\pm0.397}$& 47.1& 27.5& 88.3& 54.3$^{\pm0.266}$& 36.6& 73.2 & 54.9$^{\pm0.234}$\\

\bottomrule
\end{tabular}
\end{adjustbox}
\caption{Results (macro-avg F1) on recurrent-based language models, including TempoFormer (non-recurrent) for comparison.  \textbf{Best} scores are marked.}
\label{tab:recurrence_results}
\end{table}



\begin{table}
\centering
\begin{adjustbox}{max width=.8\linewidth}
\begin{tabular}{|l|l|l|l|}
\hline
\multicolumn{1}{|c|}{\multirow{1}{*}{\textbf{Dataset}}} &    \multicolumn{1}{c|}{\textbf{BERTScore $\downarrow$}}         & \multicolumn{1}{c|}{\textbf{Cosine Sim. $\downarrow$}}  
 & \multicolumn{1}{c|}{\textbf{Outlier $\uparrow$}}     
   \\ \hline
{LRS} & \multicolumn{1}{c|}{.457} &\multicolumn{1}{c|}{.245} & \multicolumn{1}{c|}{.867} \\ 
{TalkLife} & \multicolumn{1}{c|}{\textbf{.358} }&\multicolumn{1}{c|}{\textbf{.123}} & \multicolumn{1}{c|}{\textbf{.934}} \\ 
{Topic Shift MI} & \multicolumn{1}{c|}{.385} &\multicolumn{1}{c|}{.188} & \multicolumn{1}{c|}{.896} \\ \hline
\end{tabular}
\end{adjustbox}
\caption{Diversity Scores per Dataset.}
\label{tab:diversity}
\end{table}

\begin{table}
\begin{adjustbox}{max width=\linewidth,center}
\begin{tabular}{{|c|c|ccc|}}
\hline
 & \multicolumn{1}{|c|}{\textbf{Parameters}} & \multicolumn{3}{|c|}{\textbf{Mean Train Time/Fold} (min)} \\

\textbf{model} & 
(million) & LRS & TalkLife & Topic Shift \\\cline{1-5}

RoBERT & 
110 & 14.9& 36.0& 97.8\\

\midrule
\bottomrule

TempoFormer & 
144 & 15.6 & 38.0 & 99.1\\
\midrule
\bottomrule

RoTempoFormer & 
145 & 15.5& 37.4 & 98.9\\

\bottomrule
\end{tabular}
\end{adjustbox}
\caption{Model size and training time requirements for recurrent Transformer-based Models. Time experiments are averaged across all folds, epochs and seeds.}
\label{tab:recurrence_results_modelsize}
\end{table}

RoTempoFormer consistently outperforms RoBERT for all datasets. RoTempoFormer strikes the right balance between jointly modelling context-aware post representations and recurrence in stream dynamics. Only for LRS do recurrence-based models have a better performance than TempoFormer. To examine this phenomenon we measure the diversity of each dataset with respect to its content and report it in Table \ref{tab:diversity}. We report the BERTScore \citep{zhang2019bertscore} and Cosine similarity between SBERT pairs of representations as well as the Outlier metric \citep{larson2019outlier,stasaski2020more} on SBERT which measures the Euclidean distance between the  (unseen) posts in the test set and the mean training corpus across folds and seeds for all datasets. Thus we assess both the semantic diversity and test set diversity. Across all metrics we consistently see that TalkLife is the most and LRS the least diverse. We postulate that for more diverse datasets like TalkLife, RoBERT has a really low performance, while it performs much better on less diverse ones. This could be due to: 1) overfitting due to recurrence and 2) inability of RoBERT to jointly model diverse context-aware representations, while capturing their evolution. RoTempoFormer, maintains its high performance, striking a good balance between modelling the context-aware post-level and the timeline-level dynamics. Importantly, we thus show that TempoFormer can be used as the foundation for temporal representation learning in other architectures.

We further present the parameter and time requirements for the recurrent Transformer-based architectures (of Table \ref{tab:recurrence_results}) in Table \ref{tab:recurrence_results_modelsize}. While both TempoFormer and RoTempoFormer require around 30\% more parameters than RoBERT for training, this increase in model size is not prohibitive given the performance gains. While there is an increased model size, the overall computation requirements of less than 150M parameters are still low. Additionally, the mean training time for the TempoFormer family of models is only 1-6\% higher than for RoBERT. Time requirements across all modelas are mainly dependent on the utterance length and chosen window size for each of the datasets. 

\subsection{Model Adaptability}

\begin{table}
\centering
\begin{adjustbox}{max width=.85\linewidth}
\begin{tabular}{|l|cccc|}
\hline
 
\textbf{model} & IE & IS & O&macro-avg  \\\cline{1-5}

BERT & 43.9 & 28.1 &86.8 & 52.9  \\

RoBERTa & 46.3 &30.4 &86.6 &54.4  \\

\cline{1-5}

TempoFormer (BERT) & 50.0 &32.4 &\textbf{88.8} &57.1 \\

TempoFormer (RoBERTa) & \textbf{52.4} &\textbf{36.9} &87.3 &\textbf{58.8} \\
\hline
\end{tabular}
\end{adjustbox}
\caption{Results (macro-avg F1) on TalkLife using BERT \textit{vs} RoBERTa as the base model for TempoFormer.}
\label{tab:roberta}
\end{table}

To examine the flexibility of the TempoFormer stream-level and context-enhanced layers beyond the BERT architecture, we use TempoFormer with RoBERTa (\texttt{roberta-base}). Specifically, we allow the first 10 RoBERTa layers to model post-level (local) dynamics and modify its top two layers to capture stream dynamics. Since in Table \ref{tab:mainresults_presentation}, TalkLife benefits  from the use of RoBERTa over BERT at the post-level, we examine if this gain also transfers to the TempoFormer. Summarising results in Table \ref{tab:roberta}, we show that the RoBERTa-based TempoFormer achieves a new SOTA of 58.8\% macro-avg F1, +1.7\% over the BERT-based TempoFormer. This increase is in line with the +1.5\% performance increase between vanilla BERT and RoBERTa macro-avg F1. Importantly, the increase in overall F1 is driven by clear performance gains in the IE and IS minority classes, further demonstrating the success and adaptable nature of TempoFormer in identifying changes over time.

\section{Conclusion}

We introduce TempoFormer, a transformer-based model for change detection   operating on textual (timestamped) streams. Importantly we do so by avoiding recurrence, and only modifying the last two layers of the transformer. Furthermore, TempoFormer has the ability to model the temporal distance between textual units through a modification of rotary positional embeddings. The model achieves new SOTA, outperforming recurrent and LLM-based models on three different change detection tasks with datasets of varying temporal granularity and linguistic diversity, without loss in  generalisability. We demonstrate its usability as a foundation model in other architectures, showing it strikes the right balance between word-level, post-level and stream-level linguistic and temporal dynamics. Lastly, we showcase its flexibility in terms of base model integration, further boosting stream-level performance on par with post-level gains.


\subsection*{Limitations}

While TempoFormer shows SOTA performance on three different tasks and datasets of diverse temporal granularity involving change detection,  namely: social media overall stance shift, user mood change detection and open conversation major topic shift detection, we are yet to evaluate its performance on a wider range of tasks and datasets. Additionally, although we demonstrate strong performance in datasets as small as ~5,500 data points, we believe that our model, as most machine learning models, benefits from larger corpora in training where we can more meaningfully fine-tune the inter and intra-post relationships to model the dataset’s linguistic style and change intricacies. TempoFormer models post dynamics through a predefined stream window, identified through understanding the characteristics of a dataset via preliminary experiments. The need for initial exploration can be limiting compared to a dynamic window setting. Furthermore, despite the fact that our implementation is flexible and can be applied to different encoder architectures, the codebase is built in PyTorch, therefore imposing the constraint of PyTorch-only frameworks. On the classification front, we operate on a supervised setting therefore assuming the availability of annotated data which can be expensive to obtain especially from experts. Regarding evaluation, we focus on post-level metrics, and have not yet considered metrics more appropriate for longitudinal tasks and streams \cite{tsakalidis2022identifying}. Lastly, since our model operates by fine-tuning a pre-trained transformer-based model, like BERT, it automatically assumes the availability of such model in the language of the dataset/interest (English in our case), which might not be the case for low-resource languages.

\subsection*{Ethics Statement}

The performance of our model, TempoFormer, is demonstrated on three datasets: LRS, TalkLife and Topic Shift MI. The LRS dataset is based on the publicly available RumourEval 2017 dataset \citep{gorrell2019semeval} for stance detection, while the Topic Shift MI dataset is also a publicly available dataset based on human to human open domain conversations. Since the TalkLife dataset contains sensitive and personal user data, the appropriate Ethics approval was received from the Institutional Review Board (IRB), followed by data anonymisation and appropriate sensitive data sharing procedures. Access to this dataset was granted and approved by TalkLife \footnote{https://www.talklife.com/} through licensing for research purposes associated with the corresponding submitted proposal. All examples in the paper are paraphrased. Models were built on a secure server with authorised user-only access. The labeled TalkLife dataset and the developed models are not intended for public release in order avoid potential risks of unintended use. 

\section*{Acknowledgements}

This work was supported by a UKRI/EPSRC Turing AI Fellowship (grant no. EP/V030302/1) and Keystone grant funding from Responsible Ai UK to Maria Liakata (grant no. EP/Y009800/1), the Alan Turing Institute (grant no. EP/N510129/1), and a DeepMind PhD Scholarship to Talia Tseriotou. The work was carried out while Adam Tsakalidis was employed by Queen Mary University of London. The authors would like to thank Jenny Chim, Dimitris Gkoumas and the anonymous reviewers for their valuable feedback.


\appendix

\section{Dataset Specifics} \label{app:sec:datasets}

Since we are following 5-fold cross validation the test set consists of ~20\% of the datapoints. For LRS and Topic Shift MI the remaining data are split 25\%/75\% between dev/train sets and for TalkLife they are split 33.3\%/66.7\% between dev/train sets. The difference between these percentages is in order to ensure that we have substantial training data for LRS and Topic Shift MI in each fold as these are relatively small datasets in size.  Splitting between train/dev/test is stratified so that all timeline examples belong only to one of the sets, therefore the above percentages are approximate (not exact).

\section{Libraries}
\label{sec:app:libraries}

All experiments were ran under the same Python 3.10.12 environment including these libraries: pandas=1.5.2, matplotlib=3.7.1, pip=23.2.1, scikitlearn=1.2.0, pytorch=2.0.1, pytorch-cuda=11.8, transformers=4.35.0, tokenizers=0.14.1, huggingface-hub==0.20.3

For Seq-Sig-Net and SWNU baselines we used the Sig-Networks package and its environment as reported in \citet{tseriotou2024sig}.

\section{Computational Infrastructure}
\label{sec:app:infra}

The experiments for the LRS and TalkLife datasets were ran on a machine with 2 NVIDIA A40 GPUs of 48GB GPU RAM each, 96 cores and 256 GB of RAM.

The experiments for the Topic Shift dataset were ran on machine with 3 NVIDIA A30 GPUs of 24GB GPU RAM each, 40 cores and 384 GB	of RAM.


\section{Experimental Details}
\label{sec:app:params}



\noindent \textbf{Implementation Details} In our experiments for all models we  train on 4 epochs with early stopping and patience 3, gradient accumulation and focal loss with $\gamma=2$ and alpha of $\sqrt{1/p_{t}}$ where $p_{t}$ is the probability of class $t$ in the training data \cite{tseriotou2023sequential}. For Transformer-based models we use the AdamW optimiser \citep{loshchilov2017decoupled} and a linear scheduler and for the rest we use the Adam optimiser \citep{kingma2014adam}. The models are implemented using Pytorch \citep{paszke2019pytorch}.

For TempoFormer we use \texttt{bert-base-uncased}. We build our custom model with Huggingface's \citep{wolf2019huggingface} BERT classes and RoPE Llama classes \citep{touvron2023llama} as a starting point. All applicable BERT defaults are kept unchanged, using max length of 512 and 12 attention heads. For the classification feedforward-network we use two 64-dimensional layers and a dropout of 0.1 with ReLU. Following an initial space search, learning rate is selected using grid-search on: $[1e^{-5}, 5e^{-6}]$. \\

\noindent \textbf{BERT/RoBERTa}: Fine-tuned versions of \texttt{bert-base-uncased}/\texttt{roberta-base} using a grid search over learning rates of $\in$ $[1e^{-6}, 5e^{-6}, 1e^{-5}]$.

\noindent \textbf{FFN History}: Following \citet{tseriotou2024sig}, we perform
hyperparameter search over learning rates $\in$
$[1e^{-3}, 5e^{-4}, 1e^{-4}]$ and hidden dimensions $\in$ $[[64, 64], [128, 128], [256, 256], [512, 512]]$, over 100 epochs with a batch size of 64 and a dropout rate of 0.1.

\noindent \textbf{SWNU and Seq-Sig-Net}: We perform a hyperparameter search over: learning rates $\in$ $[0.0005, 0.0003]$, feed-forward hidden dimensions of the two layers $\in$ $[[32, 32], [128, 128]]$, LSTM hidden dimensions of SWNU units $\in$ $[10,12]$, convolution-1d reduced dimensions $\in$ $[6,10]$ and BiLSTM hidden dimensions for Seq-Sig-Net of $\in$ $[300, 400]$. Models were developed using the log-signature, time encoding in the path as well as concatenated at its output for LRS and TalkLife and sequence index in the path for Topic Shift MI. We use 100 epochs with a batch size of 64 and a dropout rate of 0.1.

\noindent \textbf{BiLSTM}: Following \citet{tseriotou2024sig}, we perform hyperparameter search over learning rates $\in$ $[1e^{-3}, 5e^{-4}, 1e^{-4}]$ and hidden
dimensions $[200, 300, 400]$, over 100 epochs with a batch size of 64 and a dropout rate of 0.1.

\noindent \textbf{SBERT}: SentenceBERT (SBERT) representations were used for different baselines \citep{reimers2019sentence} in order to obtain semantically meaningful post-level embeddings. We use 384-dimensional embeddings through \texttt{all-MiniLM-L6-v2} from the \texttt{sentence\_transformers} library.

\noindent \textbf{RoBERT}: Following \citet{pappagari2019hierarchical} we develop RoBERT with the exact same parameters as in the original paper and a grid search through learning rates $\in$ $[1e^{-6}, 5e^{-6}, 1e^{-5}]$. We follow the same grid search for RoTempoFormer.

\section{Window Results}

Full results for the window analysis are presented in Table \ref{tab:window_presentation}.

\begin{table*}[t]
\begin{adjustbox}{max width=\textwidth,center}
\begin{tabular}{{|c|ccc|cccc|ccc|}}
\hline
& \multicolumn{3}{|c|}{\textbf{LRS}}
 & \multicolumn{4}{|c|}{\textbf{TalkLife}} & \multicolumn{3}{|c|}{\textbf{Topic Shift}} \\
 
\textbf{window} & N-Sw & Sw & avg     & IE & IS & O&avg      & M & R & avg \\\cline{1-11}

\parbox[t]{2mm}{\multirow{1}{*}{5}}& 69.9&55.0& 62.5& \textbf{50.1} & 27.9 &88.7 & 55.6 & 38.3& \textbf{71.1} & 54.7\\

\midrule
\bottomrule

\parbox[t]{2mm}{\multirow{1}{*}{10}}& 73.0&56.4& 64.7& 50.0 & \textbf{32.4} &\textbf{88.8} & \textbf{57.1} & 40.3& \textbf{71.1} &55.7 \\
\midrule
\bottomrule

\parbox[t]{2mm}{\multirow{1}{*}{20}} &\textbf{75.9}& \textbf{62.0}& \textbf{68.9}& 49.5 &32.0 &\textbf{88.8} & 56.8 & \textbf{41.6}& 70.7 &\textbf{56.1} \\

\bottomrule
\end{tabular}
\end{adjustbox}
\caption{F1 scores for TempoFormer on all datasets for different window sizes. \textbf{Best} scores are marked.}
\label{tab:window_presentation}
\end{table*}

\section{Dataset Examples}\label{app:sec:examplesdataset}

Here we provide a linguistic stream example from each dataset in Tables \ref{tab:lrs_tl_example}, \ref{tab:talklife_tl_example}, \ref{tab:topicshiftmi_tl_example}.

\begin{table}
\fontsize{9.5pt}{9.5pt}\selectfont
    \caption{LRS 12-utterance long stream example with labels}
    \label{tab:lrs_tl_example}
    \begin{tabular}{  p{7.5cm}}
        \toprule
\textbf{LRS Stream}      
\\\midrule
\underline{Stream History:}\\
\textbf{U1} Approximately 50 hostages may be held captive at \#Lindt café – local reports http://t.co/1ZlzKDjvSf \#sydneysiege http://t.co/NvLr5kyQG8\\
\textbf{L1} No Switch (support)\\
\\
\textbf{U2} @RT\_com That's an exaggeration, get your facts right.\\
\textbf{L2} No Switch (deny)\\
\\
\textbf{U3} @RT\_com I thought it was only 1 from the beginning\\
\textbf{L3} No Switch (comment)\\
\\
\textbf{U4} @RT\_com 50 Hostages now\\
\textbf{L4} Switch (support)\\
\\
\textbf{U5} @RT\_com they're gonna fuck that dude up\\
\textbf{L5} Switch (comment)\\
\\
\textbf{U6} @RT\_com I pray for the safety of all the hostages; and that they are released soon.\\
\textbf{L6} Switch (comment)\\
\\
\textbf{U7} @RT\_com - "Approximately 50 hostages", in the article linked the first few lines says the number is closer to 13.\\
\textbf{L7} No Switch (deny)\\
\\
\textbf{U8} Good thing Australia has strict gun laws. "@RT\_com: Approximately 50 hostages may be held captive at \#Lindt café http://t.co/1RFsbJWl7h\\
\textbf{L8} No Switch (comment)\\
\\
\textbf{U9} @Simbad\_Reb why don't you get off Twitter and protect the next pre-school that will get hit by your infinite number of crazed gunmen\\
\textbf{L9} No Switch (comment)\\
\\
\textbf{U10} @RT\_com nah it's 5000 or maybe 500. Or Whatever sounds more alarming\\
\textbf{L10} Switch (deny)\\
\\
\textbf{U11} @RT\_com dear God!!! \\
\textbf{L11} No Switch (support)\\
\\
\underline{Current Utterance:}\\
\textbf{U12} @NijatK There is a mental health problem not a gun problem.\\
\textbf{L12} No Switch (comment)\\
\end{tabular}
\end{table}


\begin{table}
\fontsize{9.5pt}{9.5pt}\selectfont
    \caption{TalkLife 12-utterance long stream example with labels (paraphrased)}
    \label{tab:talklife_tl_example}
    \begin{tabular}{  p{7.5cm}}
        \toprule
\textbf{TalkLife Stream}      
\\\midrule
\underline{Stream History:}\\
\textbf{U1} Going to a Taylor Swift concert last week is a blessing. I feel so empowered. \\
\textbf{L1} None\\
\\
\textbf{U2} Shake it off, shake it off\\
\textbf{L2} None\\
\\
\textbf{U3} I am really craving for this feeling of getting on stage, singing my own music. It really scares me and excites me at the same time but I want to give it a chance.\\
\textbf{L3} None \\
\\
\textbf{U4} let me be brave enough to explore the unknown.
 \\
\textbf{L4} None \\
\\
\textbf{U5} he couldn't take his eyes off, what should I be thinking?\\
\textbf{L5} None \\
\\
\textbf{U6} if someone makes intense eye contact would does this mean?\\
\textbf{L6} None\\
\\
\textbf{U7} I feel the attraction but I won't do anything to hurt him. I already hurt his feelings before.\\
\textbf{L7} None \\
\\
\textbf{U8} Everyone pretends like it's not a big deal, but I can't get over the fact that I rushed my friend in the emergency room the other day. I'm deeply scarred and distressed.
\\
\textbf{L8} Switch (IS)\\
\\
\textbf{U9} I have been through so much trauma lately and I need to say it out loud that I feel broken\\
\textbf{L9} Switch (IS)\\
\\
\textbf{U10} My inspiration for singing is a burning flame, right when I thought I lost it. All these experiences helped me to rediscover music, so grateful for everything\\
\textbf{L10} None\\
\\
\textbf{U11} I'm struggling to get enough air. What's happening to me?\\
\textbf{L11} Switch (IS)\\
\\
\underline{Current Utterance:}\\
\textbf{U12} Because if you want,
I'll take you in my arms and keep you sheltered,
From all that I've done wrong\\
\textbf{L12} None \\
\end{tabular}
\end{table}


\begin{table}
\fontsize{9.5pt}{9.5pt}\selectfont
    \caption{Topic Shift MI 12-utterance long stream example with labels, denoting speakers as A and B}
    \label{tab:topicshiftmi_tl_example}
    \begin{tabular}{  p{7.5cm}}
        \toprule
\textbf{Topic Shift MI Stream}      
\\\midrule
\underline{Stream History:}\\
\textbf{U1/B} what, what do you do, now?\\
\textbf{L1} Major\\
\\
\textbf{U2/A} Well, we have saved our newspapers for years and years because the, uh, Boy Scouts our boys have been involved in have, uh, had a huge recycling bin, over at Resurrection Lutheran Church\\
\textbf{L2} Major\\
\\
\textbf{U3/B} Uh-huh.\\
\textbf{L3} Major\\
\\
\textbf{U4/A} and, uh, so we've done that for quite some time,\\
\textbf{L4} Major\\
\\
\textbf{U5/A} but since the price of paper has gone down\\
\textbf{L5} Major\\
\\
\textbf{U6/A} like it's about a fifth of what it used to be\\
\textbf{L6} Major\\
\\
\textbf{U7/B} Oh, really?
\\
\textbf{L7} Major\\
\\
\textbf{U8/A} so the Boy Scout troop quit doing it when the City took it over.\\
\textbf{L8} Major\\
\\
\textbf{U9/B} Okay.\\
\textbf{L9} Major\\
\\
\textbf{U10/A} So now we just put ours out for the City of Plano.\\
\textbf{L10} Major \\
\\
\textbf{U11/A} Do you live in Plano?\\
\textbf{L11} Rest\\
\\
\underline{Current Utterance:}\\
\textbf{U12/B} Yes,\\
\textbf{L12} Rest\\
\end{tabular}
\end{table}

\section{LLM Prompts}

To construct Mistral classification prompts we follow the recommended classification prompts as per provided guidelines \footnote{\url{https://docs.mistral.ai/guides/prompting_capabilities/\#classification}}. 
For constructing the Llama prompts we experimented with multiple prompts per dataset and identified the ones with the most stable performance. For fairer performance assessment we apply post-processing in LLM predictions to bucket them in the corresponding classification class (e.g. if the LLM generates \textit{esc} we mark it as an \textit{escalation}). In Tables \ref{tab:mistralpost}, \ref{tab:mistralstream}, \ref{tab:llamapost}, \ref{tab:llamastream} we provide our LLM prompts for the LRS dataset.

\begin{table}
    \caption{MistralInst2-7B-U for n-shot Post/Utterance-level prompting}
    \label{tab:mistralpost}
    \begin{tabular}{p{7.5cm}}
        \toprule
\textbf{MistralInst2-7B-U Template}      
 
\\\midrule
You are a helpful, respectful and honest assistant for labeling online Twitter conversations between users.
Given the online post of a user in a conversation stream around a rumourous claim on a newsworthy event which it is discussed by tweets in the stream, determine if in the current post there is a switch with respect to the overall stance.
Answer with "none" for either the absence of a switch or cases where the numbers of supporting and opposing posts are equal and with "switch" for switch between the total number of oppositions (querying or denying) and supports or vice versa.
Your task is to assess and categorize post input after $<<<>>>$ into one of the following predefined outputs:\\

     \\
none\\
switch \\
\\
You will only respond with the output. Do not include the word "Output". Do not provide explanations or notes.\\
\\
\#\#\#\#\\
Here are some examples:\\
\\
Input: \textit{post example 1}\\
Output: \textit{post label 1}\\
$\cdots$\\
Input: \textit{post example n}\\
Output: \textit{post label n}\\
\#\#\#\#\\
\\
        \bottomrule
    \end{tabular}
\end{table}

\begin{table}
    \caption{MistralInst2-7B-S for n-shot Stream-level prompting}
    \label{tab:mistralstream}
    \begin{tabular}{  p{7.5cm}}
        \toprule
\textbf{MistralInst2-7B-S Template}      
 
\\\midrule
You are a helpful, respectful and honest assistant for labeling online Twitter conversations between users.
Given the most recent online conversation history between users around a rumourous claim on a newsworthy event, determine if the most recent input user post is a switch with respect to the overall conversation stance.
Answer with "none" for either the absence of a switch or cases where the numbers of supporting and opposing posts are equal and with "switch" for switch between the total number of oppositions (querying or denying) and supports or vice versa.
Your task is to assess and categorize post input after $<<<>>>$ into one of the following predefined outputs:\\

     \\
none\\
switch \\
\\
You will only respond with the output. Do not include the word "Output". Do not provide explanations or notes.\\
\\
\#\#\#\#\\
Here are some examples:\\
\\
Conversation History:\\
$u_{a-4}$\\
$u_{a-3}$\\
$u_{a-2}$\\
$u_{a-1}$\\
Input: \textit{post example 1, $u_a$}\\
Output: \textit{post label 1}\\
$\cdots$\\
Conversation History:\\
$u_{b-4}$\\
$u_{b-3}$\\
$u_{b-2}$\\
$u_{b-1}$\\
Input: \textit{post example n, $u_b$}\\
Output: \textit{post label n}\\
\#\#\#\#\\
\\
        \bottomrule
    \end{tabular}
\end{table}

\begin{table}
    \caption{Llama2-7B-U for n-shot Post/Utterance-level prompting}
    \label{tab:llamapost}
    \begin{tabular}{  p{7.5cm}}
        \toprule
\textbf{Llama2-7B-U Template}      
 
\\\midrule
$<s>[INST] <<SYS>>$\\
You are a helpful, respectful and honest assistant for labeling online Twitter conversations between users.\\
$<</SYS>>$\\
\\
Given the online post of a user in a conversation stream around a rumourous claim on a newsworthy event which it is discussed by tweets in the stream, determine if in the current post there is a switch with respect to the overall stance.
\\
Answer with "none" for either the absence of a switch or cases where the numbers of supporting and opposing posts are equal and with "switch" for switch between the total number of oppositions (querying or denying) and supports or vice versa. \\
\\
Example 1:\\
Input: \textit{post example 1}\\
Output: \textit{post label 1}\\
$\cdots$\\
Example n:\\
Input: \textit{post example n}\\
Output: \textit{post label n}\\
\\
Only return "none" or "switch".\\ 
Limit the answer to 1 word.\\
$[/INST]$\\
$</s>$\\
\\
        \bottomrule
    \end{tabular}
\end{table}

\begin{table}
    \caption{Llama2-7B-S for n-shot Stream-level prompting}
    \label{tab:llamastream}
    \begin{tabular}{  p{7.5cm}}
        \toprule
\textbf{Llama2-7B-S Template}      
 
\\\midrule
$<s>[INST] <<SYS>>$\\
You are a helpful, respectful and honest assistant for labeling online Twitter conversations between users.\\
$<</SYS>>$\\
\\
Given the most recent online conversation history between users around a rumourous claim on a newsworthy event, determine if the most recent input user post is a switch with respect to the overall conversation stance.
\\
Answer with "none" for either the absence of a switch or cases where the numbers of supporting and opposing posts are equal and with "switch" for switch between the total number of oppositions (querying or denying) and supports or vice versa.\\
\\
Example 1:\\
Conversation History:\\
$u_{a-4}$\\
$u_{a-3}$\\
$u_{a-2}$\\
$u_{a-1}$\\
Input: \textit{post example 1, $u_{a}$}\\
Output: \textit{post label 1}\\
$\cdots$\\
Example n:\\
Conversation History:\\
$u_{b-4}$\\
$u_{b-3}$\\
$u_{b-2}$\\
$u_{b-1}$\\
Input: \textit{post example n, $u_{b}$}\\
Output: \textit{post label n}\\
\\
Only return "none" or "switch".\\ 
Limit the answer to 1 word.\\
$[/INST]$\\
$</s>$\\
\\
        \bottomrule
    \end{tabular}
\end{table}

\label{sec:appendix}

\end{document}